\documentclass{article}
\usepackage{spconf,amsmath,graphicx}
\usepackage{multirow}
\usepackage{makecell}
\usepackage{color}

\title{GRASS: UNIFIED GENERATION MODEL FOR SPEECH-TO-SEMANTIC TASKS}

\name{Aobo Xia$^{\star \dagger}$\thanks{Work done while the first author was an intern at Didi Chuxing.} \qquad ShuYu Lei$^{\dagger}$\thanks{Corresponding author is the second author.} \qquad YuShu Yang$^{\dagger}$ \qquad Xiang Guo$^{\dagger}$ \qquad Hua Chai$^{\dagger}$}
  
  \address{$^{\star}$ Beijing University of Posts and Telecommunications, Beijing, China \\
      $^{\dagger}$ Didi Chuxing, Beijing, China\\
      {\small \tt aoboxia@bupt.edu.cn, \{leishuyu, yangyushu, guoxiang, chaihua\}@didiglobal.com}}

\begin{document}
%\ninept
%
\maketitle
\begin{abstract}
This paper explores the instruction fine-tuning technique for speech-to-semantic tasks by introducing a unified end-to-end (E2E) framework that generates target text conditioned on a task-related prompt for audio data. We pre-train the model using large and diverse data, where instruction-speech pairs are constructed via a text-to-speech (TTS) system. Extensive experiments demonstrate that our proposed model achieves state-of-the-art (SOTA) results on many benchmarks covering speech named entity recognition, speech sentiment analysis, speech question answering, and more, after fine-tuning.
Furthermore, the proposed model achieves competitive performance in zero-shot and few-shot scenarios. To facilitate future work on instruction fine-tuning for speech-to-semantic tasks, we release our instruction dataset and code.
\end{abstract}


\section{Introduction}
\label{sec:introduction}
speech-to-semantic comprises a wide range of diverse tasks, such as speech intent detection, speech sentiment analysis, and speech question answering, which play a crucial role in many applications.

Traditionally, a standard pipeline fashion transcribes audio into text using automatic speech recognition (ASR) and then maps the transcribed text to a semantic structure using natural language understanding (NLU). 
This pipeline approach has two limitations: 1) transcription errors are caused by ASR, and 2) the absence of acoustic information, such as stress and intonation, is not considered in NLU.
% This pipeline solution suffers from the error transcript caused by ASR and the lack of acoustic information, such as stress and intonation, utilized in NLU. 

Thus, there has been an active area of E2E methods \cite{chen2018spoken,serdyuk2018towards,haghani2018audio} to infer semantic labels directly from audio data.
Although the above E2E approaches achieve promising results, there is still a major limitation of the requirement of a large amount of label data. 
To alleviate this limitation, prior research \cite{huang2020leveraging,lugosch2020using,mdhaffar2022end,pasad2022use} has utilized the data augmentation method to generate speech-to-semantic pairs.
However, the aforementioned studies use architectures specifically designed for their task.
Consequently, it is brittle and sensitive to slight changes in the data distribution and task specific.
This motivates previous research \cite{radford2019language} using a unified model to handle various text-to-text tasks and achieving promising results.

A nascent line of research \cite{wei2021finetuned,sanh2022multitask,chung2022scaling}, named instruction fine-tuning, has been shown to improve model performance and generalize to unseen tasks for the language model.
These language models are fine-tuned to learn the alignment between various instructions and results.
Therefore, they have the ability to handle various text tasks.

Inspired by the aforementioned methods, the goal of this paper is to investigate a unified generation model to handle various speech-to-semantic tasks via instruction fine-tuning.
We propose a new speech-to-semantic model called \textbf{GRASS}, which stands for unified \textbf{G}ene\textbf{RA}tion model for \textbf{S}peech to \textbf{S}emantic.
% This paper investigates a unified generation model to infer semantic labels conditioned on a task-related prompt for audio data.
GRASS infers target text conditioned on a task-related prompt for audio data, which is based on an architecture of the Whisper model \cite{pmlr-v202-radford23a} and is pre-trained by instruction-speech pairs.
However, this approach faces a significant challenge due to the scarcity of annotated speech-to-semantic data. 
In the related field of natural language processing, there is an abundance of instruction datasets available in text-to-text format, which can be converted into speech-to-semantic data by TTS.
% 简写版本
Although this is an extreme concern about the disparity between TTS-generated speech data and real speech,
it is important to emphasize that our primary objective is to enable the model to align the audio data with semantic results conditioned on the given instructions.
% 详细写作版本

As a result, the gap between TTS-generated data and real speech is not significant. 
In practice, we selected a subset from the Super-NaturalInstructions \cite{wang2022super} and Stanford-Alpaca \cite{alpaca} datasets that is applicable to the speech-to-semantic tasks and generated data via a TTS interface on Microsoft Azure\footnote{https://azure.microsoft.com/en-us/products/ai-services/text-to-speech}.
During the TTS process, different voice options, such as genders, countries, and speakers, are randomly set to generate diversity audio data.

\begin{figure*}[t]
\centering
\includegraphics[width=0.94 \textwidth]{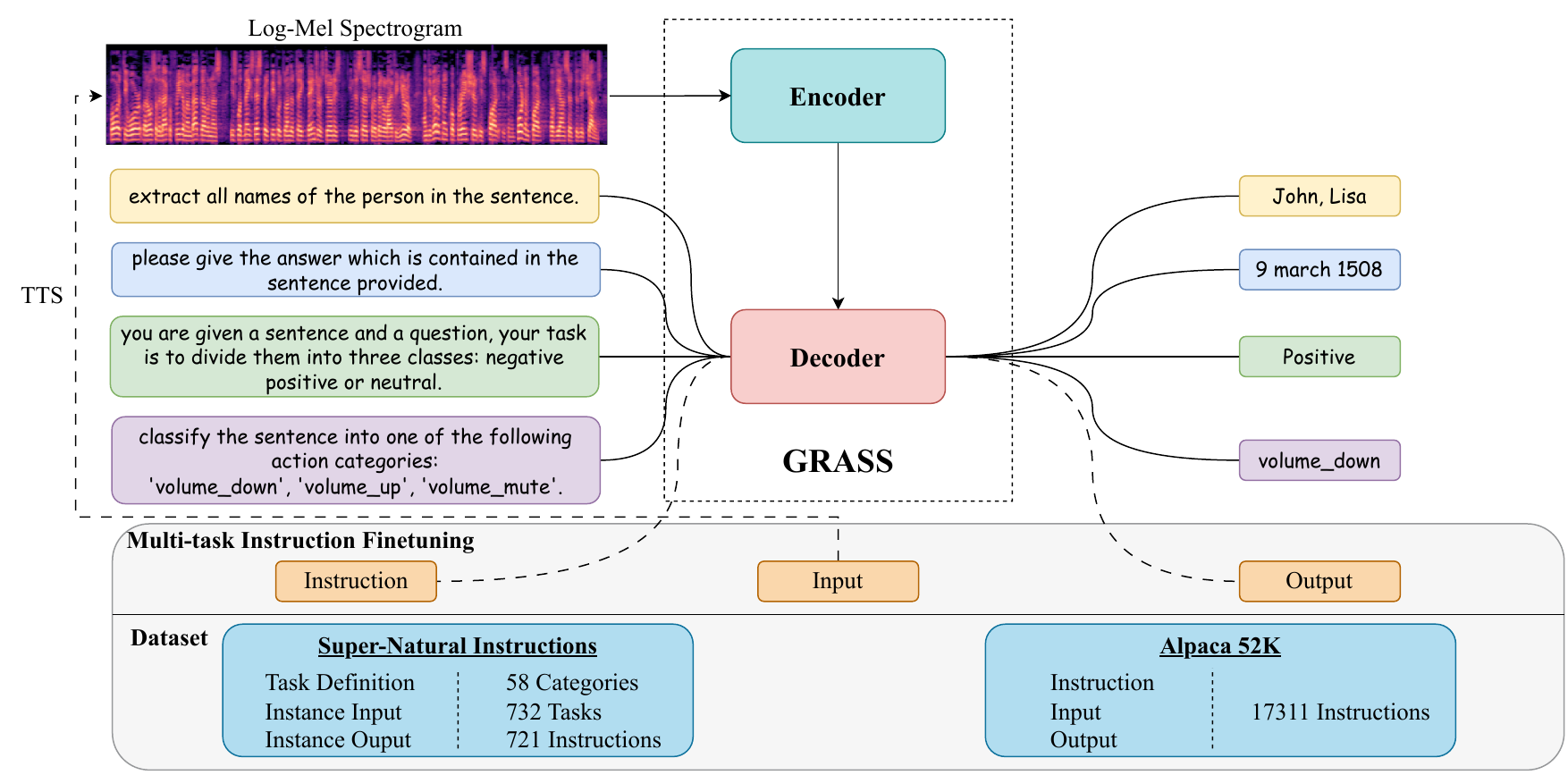}
\caption{A diagram of GRASS framework. We fine-tune a sequence-to-sequence Transformer model on diverse pairs of instruction and audio data generated through TTS across various tasks, where target text is generated from decoder on conditioning instruction text for audio embedding. Solid lines represent the inference stage, while dashed lines represent the instruction fine-tuning stage.}
% where target text is generated from decoder on conditioning instruction text for audio embedding achieved by encoder.
% A sequence composed of instruction text embendding and audio embbedding  outputted by the encoder is cast as feeding decoder as input for predicting the target text.}
\label{fig:model}
% \vspace{-0.5cm}
\end{figure*}
Extensive experiments demonstrate that GRASS significantly outperforms SOTA models after fine-tuning in all downstream tasks.
Furthermore, GRASS achieves competitive performance in part of tasks after a few-shot fine-tuning.

Our contributions are summarized as follows:
\begin{itemize}
    \item We propose a unified generation model called GRASS to handle various task on speech-to-semantic, where GRASS encodes audio data and decoder target text conditioned on a task-related prompt. 
    \item We utilize instruction fine-tuning to access the GRASS achieved new SOTA on all speech-to-semantic tasks.
    % \item We release our code and created speech instruction data to encourage further research.
    \item We release our code and constructed data to encourage further research\footnote{https://github.com/aoboxia/Grass}.
\end{itemize}

\section{Related work}
\label{sec:relatedwork}

% \subsection{E2E speech semantic understanding}
Recent studies \cite{chen2018spoken,serdyuk2018towards,haghani2018audio} build an E2E speech-to-semantic model to avoid the shortcomings of the traditional pipeline method, which requires a large amount of speech-to-semantic data.
% propagation of ASR errors and to consider acoustic information for semantic understanding, 
This is a challenging situation that often occurs during the development of new applications.
Thus, several studies \cite{huang2020leveraging,lugosch2020using,mdhaffar2022end,pasad2022use} have used the data augmentation method to generate speech-to-semantic pairs, such as converting text-semantic data into speech-to-semantic pairs with TTS \cite{huang2020leveraging,lugosch2020using}, predicting pseudolabels for the text of speech-to-text pairs with NLU \cite{mdhaffar2022end}, and aligning acoustic and text embeddings \cite{pasad2022use}.
The above studies use task-specific architectures that are sensitive to data distribution.
% \cite{huang2020leveraging,lugosch2020using} uses TTS to convert text-semantic data to speech-to-semantic pairs.
% \cite{mdhaffar2022end} uses trained NLU to predict pseudolabels for the text of speech-to-text pairs to generate speech-to-semantic data.
% \cite{pasad2022use} maps the text of all text-semantics pairs to speech embeddings to create pseudospeech-to-semantics pairs.
% Furthermore, transfer learning is utilized to improve the target semantic task in the study \cite{tomashenko2019investigating}.
Instead, our work considers a unified model to handle various speech-to-semantic tasks, which may benefit from using instruction fine-tuning.
% Instead, our work utilizes generated speech instruction data to fine-tune the Whisper model without needing a great deal of semantically labeled data in few-shot or zero-shot in speech semantic understanding.

% \subsection{instruction fine-tuning}
A nascent line of research \cite{wei2021finetuned,sanh2022multitask,chung2022scaling}, named instruction fine-tuning, fine-tunes a pre-trained model with instructions to improve its performance and generalization to unseen tasks in text-to-text style tasks.
% \cite{wei2021finetuned} first explores the instruction fine-tuning method to improve the performance of language models in unseen tasks, they evaluate trained language models with a single held-out task. 
% Compared to prior studies, \cite{sanh2022multitask} evaluates the model's ability of generalization by multiple held-out tasks.
% Furthermore, \cite{chung2022scaling} extend prior research by scaling the number of tasks, scaling the model size, and fine-tuning on chain-of-thought data.
Inspired by instruction fine-tuning, our work uses instruction-speech pairs to pre-train GRASS to access better performance on various speech-to-semantic tasks.

% \subsection{Speech representation learning}
Speech representation learning \cite{pmlr-v202-radford23a,baevski2021unsupervised,hsu2021hubert} has produced promising results in speech recognition.
In particular, the Whisper model \cite{pmlr-v202-radford23a} is useful for performing multitasks, such as multilingual speech recognition, speech translation, and language identification.
The whisper model uses task-specific tokens to generate results for above aligning speech-to-text tasks, which is not applied directly to speech-to-semantic tasks.
% In contrast to the Whisper model, our proposed model is fine-tuned with constructed speech-to-text instruction data, which contain div, aiming to achieve a unified generation model for various speech-to-semantic tasks.
In contrast to the Whisper model, GRASS is pre-trained with constructed instruction-speech data, and the task-related instruction is phrased in natural language instead of task-specific tokens, aiming to achieve a unified generation model for various speech-to-semantic tasks.

\begin{table*}[t]
    \centering
    \caption{The overall results on various speech semantic tasks.}
    \vspace{0.05in}
    \resizebox{0.85\textwidth}{!}{
    \begin{tabular}{ccccccccc}
    \hline
        \multirow{3}{*}{\textbf{Model}} & \multirow{3}{*}{\textbf{Data Size}} & SLUE-NER & SLUE-SA & \multicolumn{2}{c}{SLUE-QA} & FSC &  \multicolumn{2}{c}{SLURP}\\
        ~ & ~ & NER & SA & \multicolumn{2}{c}{QA} & DAC & IC & SF \\ 
        ~ & ~ & Micro F1 & Macro F1 & BLEU-4 & Acc & Acc & Acc & Slu F1  \\ \hline
        Baselines & 100\% & 70.3 & 45.3 & - & - & 99.71 %(espnet-slu)
        & 86.52 %(speechbrain) 
        & 76.91 \\ %(espnet-slu)  \\ 
        GRASS & 1\% & 27.27 & 50.47 & 11.17 & 26.07 & 91.24 & 54.62 & 45.72  \\ 
        GRASS & 5\% & 55.53 & 52.9 & 12.92 & 31.33 & 99.6 & 75.74 & 63.67  \\ 
        % GRASS & 25\% & \textcolor{red}{71.15} & \textcolor{red}{54.89} & 16.53 & 40.11 & \textcolor{red}{99.71} & \textcolor{red}{87.23} & 74.83  \\ 
        GRASS & 25\% & 71.15 & 54.89 & 16.53 & 40.11 & 99.71 & 87.23 & 74.83  \\ 
        GRASS & 100\% & \textbf{74.21} & \textbf{57.02} & \textbf{30.37} & \textbf{47.28} & \textbf{99.76} & \textbf{88.17} & \textbf{77.47} \\
        \hline
        	
        % \makecell[l]{Whisper\\ -w/o instruction} & 100\% & 70.93 & 53.95 & 28.87 & 41.7 & 99.73 & 87.16 & 74.85  \\ \hline
        Whisper Large V2 & 100\% & 70.93 & 53.95 & 28.87 & 41.7 & 99.73 & 87.16 & 74.85  \\ \hline
    \end{tabular}}
    \label{tab:result}
\end{table*}
\section{Methodology}
\label{sec:methodology}
We show the overall framework of GRASS in Figure \ref{fig:model}, which is mainly consists of two parts: a unified generation model and data collection.
We detail the two parts in the following subsections.
 
\subsection{Unified Generation Model}

We now formally define the speech-to-semantic task notions for a unified generation model.
Given a sequence of audio feature vectors $X = [x_1, \dots, x_l]$ and a sequence of instruction text tokens $I=[i_1, \dots, i_m]$ defined in natural language, the goal of GRASS is to predict a sequence of target text  $Y=[y_1, \dots, y_n]$ as
\begin{equation}
    Y = \arg\max_{Y}P(Y|X,I).
\end{equation}

To this end, we use the architecture of the Whisper model to generate semantic labels $Y$ conditioned on audio feature vectors $X$ and instruction text tokens $I$, as illustrated in Figure \ref{fig:model}.
We only train the model to predict the sequence of semantic labels where the training losses on the instruction tokens are masked out.
We will introduce the details of the implementation in Section \ref{subsec:implement}.

\subsection{Data collection}
Instruction fine-tuning \cite{wei2021finetuned} has been shown to improve model performance and generalize to unseen tasks in text-to-text tasks.
Inspired by this, we explore the instruction fine-tuning method for the speech-to-text style to generalize on various speech semantic tasks.
To the best of our knowledge, there is no large number of pair of instruction-speech data available.
Thus, we collect instruction-speech pairs, which are constructed on the basis of two text instruction datasets: Super-NaturalInstructions \cite{wang2022super} and Stanford-Alpaca \cite{alpaca}. we keep the original instruction and target text and generate the speech data for the input text via a TTS system.
Despite extreme concern about the disparity between TTS-generated speech data and real speech, it is worth noting that generated speech has reached a level of approximation close to natural speech, which benefits from the recent TTS technology.
% This is partly due to the remarkable advances in TTS technology, which have made this inexpensive and efficient method of data collection feasible. 

We create instruction data mainly based on the Super-NaturalInstructions dataset.
For the diversity of instruction, we also extend fine-tuning data with Stanford-Alpaca dataset.
For a convenient transformation to a speech-to-text style, we convert the Super-NaturalInstructions dataset into the Stanford-Alpaca format, which consists of instruction, input, and output.
The instruction describes the task that the model should perform.
The input is the instance for the task that the model should handle.
The output is the answer to the instruction based on the input.
During converting the Super-NaturalInstructions dataset, we treat task definition as instruction and convert instance into input and output.
% We discard the positive and negative examples since these text examples cannot be treated as speech-to-text examples for in-context learning in GRASS.
In addition, we filter some irrelevant tasks in the above datasets, which are essential different from speech tasks, such as spam classification, spelling error detection, and fill-in-the-blank.
We also remove examples where the input is not suitable for phonetic expressions, including URLs, codes, and special characters.

After unifying the format and filtering the examples, for each example, we transform the input text into speech to generate a fine-tuning instance: constructed audio as input of the encoder, instruction text as input of the decoder, and output text as target of the decoder as shown in Figure \ref{fig:model}.
To transform the input text, we use the Microsoft Azure TTS interface, which is readily available.
During the TTS process, we randomly selected different genders, 14 different countries, and 79 pre-set speakers to generate diversity audio data.
Finally, we obtain 391,864 instruction-speech instances from Super-NaturalInstructions and 17,313 instruction-speech instances from Stanford-Alpaca, respectively.
% Finally, we present the detailed statistics of the constructed data in Table \ref{tab:data}.

% \begin{table}[htbp]
%   \centering
%   \caption{Statistics of the constructed data.}
  
%   \resizebox{0.45\textwidth}{!}{% <-
%     \begin{tabular}{ccc}
%     \hline
%     dataset & data size & instruction size \\
%     \hline
%     Super-NaturalInstructions  & 391,864 & 721 \\
%     Stanford-Alpaca & 17,313 & 17,313 \\
%     \hline
%     \end{tabular}%
%     }
%   \label{tab:data}%
  
% \end{table}%

\begin{table*}[t]
    \centering
    \caption{The overall results in zero-shot settings.}
    \vspace{0.05in}
    \resizebox{0.96\textwidth}{!}{
    \begin{tabular}{cccccccccc}
    \hline
        Number of & SLUE-NER & SLUE-SA & \multicolumn{2}{c}{SLUE-QA} & \multicolumn{3}{c}{FSC} & \multicolumn{2}{c}{SLURP} \\
        instruction & Micro F1 & Macro F1 & Bleu-4 & Acc & action(Acc) & object(Acc) & location(Acc) & scenario(Acc) & action(Acc)\\ \hline
        721 & 16.73 & 42.02 & 9.7 & 16.82 & \textbf{27.31} & 30.97 & 53.15 & 11.76 & 22.8 \\ \hline
        4285 & \textbf{18.62} & \textbf{49.04} & \textbf{14.3} & \textbf{19.37} & 23.83 & \textbf{51.83} & \textbf{56.05} & \textbf{12.8} & \textbf{33.07} \\ \hline
    \end{tabular}}
    \label{tab:zeroshot}
\end{table*}
\section{Experiments}
\label{sec:experiments}
\subsection{Implementation details}
For the instruction fine-tuning process, we chose a `large-v2' version of Whisper as the base model to obtain GRASS.
To save computing resources, we only fine-tune the parameters of the decoder part.
For the hyperparameters, we set the dropout rate at 0.1, the batch size at 16, and the learning rate at 1e-4.
We update the model parameters after accumulating 4 gradients.
We employ the same loss function and optimizer as those utilized in the Whisper model.
% To simulate a bigger batch size, we set the accumulated steps at 4.
During the training stage, we save a checkpoint every 5000 steps.
We chose the best checkpoint based on the evaluation results of the downstream tasks.

During the fine-tuning stage in downstream tasks, we search the learning rate from the range [1e-5, 3e-5, 1e-4] and the batch size from [4, 8].
Meanwhile, we augment the waveforms of every data with a speed perturbation by a factor of 0.95 and 1.05.
We implement our code based on the speechbrain toolkit\footnote{https://speechbrain.github.io/}.
Finally, we report the best results.
\label{subsec:implement}

\subsection{Evaluation datasets and baselines}
In this section, we present GRASS fine-tuning results on 5 speech-to-semantic tasks: SLUE-NER, SLUE-SA, SLUE-QA, FSC, SLURP.

\textbf{SLUE-NER}\cite{shon2022slue}: a speech named entity recognition (NER) involves detecting the named entities and their tags (types) in a given speech.
We evaluate the performance of SLUE-NER using micro-averaged (Micro) F1 scores. In this work, we follow the original paper results on the dev set, as the test set is not publicly accessible.
% We use the SOTA result as a report in .

\textbf{SLUE-SA}\cite{shon2022slue}: a sentiment analysis (SA) refers to classifying a given speech segment as having negative, neutral, or positive sentiment. We evaluate the performance of SLUE-SA using macro-averaged (Macro) F1 scores. In this work, we follow the original paper results on the dev set, as the test set is not publicly accessible.

\textbf{SLUE-QA}\cite{shon2022slue2}: a question answering (QA) infers the answer in a given speech. 
Since the proposed GRASS is an E2E solution for question answering, we evaluate the performance using BLUE-4 and accuracy instead of the original frame f1 scores. We concatenate original document with question audio as input, which is truncated to 30 seconds, and then output the answer.

\textbf{FSC}\cite{lugosch2019speech}: a dialogue action classification (DAC) predicts the `action', `object', and `location' labels of a given speech.
We evaluate the performance of the FSC using accuracy and we include results from the SOTA model as reported in \cite{qian2021speech} for comparison.

\textbf{SLURP}\cite{bastianelli2020slurp}: a intent classification (IC) and a slot filling (SF) are introduced, where action and entities are also called `intent' and the entities consist of `slots' and `values'.
We evaluate the performance of intent classification using accuracy and slot filling using SLU-F1 as proposed in \cite{bastianelli2020slurp}, respectively. We include results from the SOTA model from speechbrain for comparison.

\subsection{Overall results}

We illustrate the results of employing different fine-tuning data sizes comparing current SOTA in the E2E fashion, as shown in Table \ref{tab:result}.
We observe that GRASS fine-tuning with the 100\% data size significantly outperforms SOTA.
Considering the effectiveness of instruction fine-tuning, we also compare fine-tuning GRASS with fine-tuning raw Whisper.
As shown in Table \ref{tab:result}, fine-tuning GRASS  significantly outperforms fine-tuning raw Whisper. 

The above results indicate that GRASS can significantly improve various speech-semantic task performance by instruction fine-tuning.
% On the NER task, the improvement was 4.4\%, on the SA task, the improvement was 5.7\%, on the QA task, the BLEU and ACC scores were improved by 5.2\% and 11.8\%, respectively, and on the IC/SF task, the intent accuracy and SLU F1 scores were improved by 1.6\% and 2.7\%, respectively.

In order to verify the robustness of our method and its effectiveness in a low-resource scenario, we randomly sample 1\%, 5\% and 25\% of the training data from the downstream task and evaluate the performance of different variants of the model when fine-tuned on the sampled data.
As shown in Table \ref{tab:result}, GRASS fine-tuning with only 25\% training data outperforms SOTA on each task, except for the SF task.
% Since there are 56 categories of slots in the SF task, the requirement of the SF task for training data is very large.
% For SF tasks, there is still a certain distance from SOTA, since there are 56 slot categories in the SLURP dataset, and the SLU-F1 score depends on the word and character accuracy of entity prediction. 
% Therefore, the requirement for training data is very large, and the long tail problem faced by a small amount of data is more serious.
Fine-tuning GRASS with only 1\% of the training data (58 examples) on SA task outperforms SOTA.
Meanwhile, fine-tuning GRASS with only 5\% of the training data on the DAC task only drops 0.4\% in accuracy.

This result indicates that our method can also achieve good results in the downstream task in the low-resource scenario.

% \begin{table*}[t]
%     \centering
%     \caption{The overall results on various speech semantic tasks.}
%     \vspace{0.05in}
%     \resizebox{0.65\textwidth}{!}{
%     \begin{tabular}{lcccccccc}
%     \hline
%         \multirow{3}{*}{\textbf{Model}} & \multirow{3}{*}{\textbf{Data Size}} & SLUE-NER & SLUE-SA & FSC &  \multicolumn{2}{c}{SLURP}\\
%         ~ & ~ & ner & sa & dac & intent & sf  \\ 
%         ~ & ~ & Micro F1 & Macro F1  & Acc & Acc & Slu F1  \\ \hline
%         baseline & 100\% & 70.3 & 45.3 & 99.71 %(espnet-slu)
%         & 86.52 %(speechbrain) 
%         & 76.91 \\ %(espnet-slu)  \\ 
%         whisper-large & 1\% & 27.27 & 50.47 & 91.24 & 54.62 & 45.72  \\ 
%         whisper-large & 5\% & 55.53 & 52.9 & 99.6 & 75.74 & 63.67  \\ 
%         whisper-large & 25\% & \textcolor{red}{71.15} & \textcolor{red}{54.89} & \textcolor{red}{99.71} & \textcolor{red}{87.23} & 74.83  \\ 
%         whisper-large & 100\% & \textbf{74.21} & \textbf{57.02} & \textbf{99.76} & \textbf{88.58} & \textbf{76.88} \\
%         \hline
%         \makecell[l]{whisper-large\\ -w/o instruction} & 100\% & 70.93 & 53.95 & 99.73 & 87.16 & 74.85  \\ \hline
%     \end{tabular}}
%     \label{tab:result}
% \end{table*}

\subsection{Zero-shot ability}
We explore GRASS for the zero-shot ability and rephrase the instruction through GPT-4 to diversify the task-related prompt.
We detail the above two zero-shot experiments in Table \ref{tab:zeroshot}.
We observe that expanding instruction significantly improves performance in zero-shot settings.
Therefore, it is necessary to enhance the diversity of instructions for the zero-shot scenario, which indicates a promising direction for future work.

% \subsection{Visualization}

\section{Conclusion}
\label{sec:conclusion}
In this paper, we first propose a unified generation model for various speech-to-semantic tasks called GRASS.
To the best of our knowledge, our work is the first successful attempt at utilizing instruction fine-tuning for speech-to-semantic tasks by evaluating on fine-tuning downstream tasks.
Second, we collect the instruction data on speech by accessible TTS for instruction fine-tuning and open source them to facilitate future work.
Third, we also carried out experiments in low-resource scenarios that demonstrate that GRASS achieves competitive performance, which alleviates the requirement for a large amount of labeled data.

% \newpage
% \fontsize{9.9pt}{10.0pt}
% \selectfont
\bibliographystyle{IEEEbib}
\bibliography{strings,refs}

\end{document}